\documentclass[conference]{IEEEtran}
\IEEEoverridecommandlockouts

\usepackage{cite}
\usepackage{amsmath,amssymb,amsfonts}
\usepackage{graphicx}
\usepackage{textcomp}
\usepackage{xcolor}
\usepackage{multirow}
\usepackage{booktabs}
\usepackage{hyperref}
\usepackage{tikz}
\usepackage{comment}
\usepackage{algpseudocode}
\usepackage{authblk}
\usepackage{subcaption}
\usepackage{threeparttable}
\usepackage{bbm}
\usepackage[font=small,labelfont=bf]{caption}
\usepackage{amssymb}
\usepackage{pifont}

\definecolor{ao(english)}{rgb}{0.0, 0.5, 0.0}

\newcommand{\green}[1]{\textcolor{ao(english)}{#1}} 
 
\newcommand{\blue}[1]{\textcolor{blue}{#1}} 
 
\newcommand{\cmark}{\ding{51}}%
\newcommand{\xmark}{\ding{55}}%
\hypersetup{
    colorlinks=true,
    linkcolor=blue,
    filecolor=magenta,      
    urlcolor=cyan,
    citecolor = magenta,
    pdfpagemode=FullScreen,
}

\makeatletter
\newcommand*\titleheader[1]{\gdef\@titleheader{#1}}
\AtBeginDocument{%
  \let\st@red@title\@title
  \def\@title{%
    \bgroup\normalfont\large\centering
    \vskip-3.9em
    \@titleheader\par\egroup
    \vskip0.8em\st@red@title}
}
\makeatother

\def\BibTeX{{\rm B\kern-.05em{\sc i\kern-.025em b}\kern-.08em
    T\kern-.1667em\lower.7ex\hbox{E}\kern-.125emX}}

\titleheader{\vspace{10pt} \normalsize{2025 IEEE International Symposium on Performance Analysis of Systems and Software (ISPASS)}}

\title{Generative AI in Embodied Systems: System-Level Analysis of Performance, Efficiency and Scalability} \vspace{-20pt}

\author[$1$]{Zishen~Wan$^{*}$\thanks{$^{*}$Corresponding Email: zishenwan@gatech.edu}}
\author[$1$]{Jiayi Qian}
\author[$2$]{Yuhang Du}
\author[$3$]{Jason Jabbour}
\author[$3$]{Yilun Du}
\author[$2$]{Yang (Katie) Zhao}
\author[$1$]{\\Arijit Raychowdhury}
\author[$1$]{Tushar Krishna}
\author[$3$]{Vijay Janapa Reddi}

\affil[$ $]{\normalsize \textit{$^{1}$Georgia Institute of Technology, GA \hspace{0.1in} $^{2}$University of Minnesota, Twin Cities, MN\hspace{0.1in} $^{3}$Harvard University, MA}}

\begin{document}

\maketitle

\begin{abstract}
Embodied systems, where generative autonomous agents engage with the physical world through integrated perception, cognition, action, and advanced reasoning powered by large language models (LLMs), hold immense potential for addressing complex, long-horizon, multi-objective tasks in real-world environments. However, deploying these systems remains challenging due to prolonged runtime latency, limited scalability, and heightened sensitivity, leading to significant system inefficiencies.
In this paper, we aim to understand the workload characteristics of embodied agent systems and explore optimization solutions. We systematically categorize these systems into four paradigms and conduct benchmarking studies to evaluate their task performance and system efficiency across various modules, agent scales, and embodied tasks. Our benchmarking studies uncover critical challenges, such as prolonged planning and communication latency, redundant agent interactions, complex low-level control mechanisms, memory inconsistencies, exploding prompt lengths, sensitivity to self-correction and execution, sharp declines in success rates, and reduced collaboration efficiency as agent numbers increase.
Leveraging these profiling insights, we suggest system optimization strategies to improve the performance, efficiency, and scalability of embodied agents across different paradigms. This paper presents the first system-level analysis of embodied AI agents, and explores opportunities for advancing future embodied system design.
\end{abstract}

\section{Introduction}
\label{sec:intro}

Embodied AI systems represent intelligent agents that interact with the physical world through perception, cognition, and action to perform complex tasks~\cite{duan2022survey,wan2024thinking,xu2024survey,liu2024aligning,wan2025reca}. These systems integrate advanced cognitive frameworks with environmental sensing and task execution to navigate long-horizon multi-objective challenges, such as motion planning and autonomous decision making. Large language models (LLMs) have been increasingly incorporated into embodied AI systems to enhance planning, communication, and reasoning capabilities in dynamic and uncertain environments. These systems combine high-level reasoning from LLMs with precise low-level execution to enable robust and adaptive behaviors in complex tasks that require continuous engagement with the environment.

Embodied AI agents hold significant potential for real-world applications, including robotics, autonomous vehicles, and collaborative multi-agent systems. These systems have demonstrated superior capabilities in long-horizon multiobjective tasks across various domains. For example, CoELA~\cite{zhang2024building} improves the success rate by 23\% in complex object transport, table setting, and human-agent interaction tasks. The ability of embodied AI systems to adapt to dynamic environments, process multimodal input, and execute complex action sequences makes them well-suited for tasks that require continuous interaction with the real world.

Despite their promising application potential, embodied AI systems face critical challenges. First, the \emph{high computational latency} in long-horizon tasks arises from repeated inference runs and complex planning processes. Second, \emph{system inefficiencies}, such as redundant communication and sequential processing, lead to increased runtime and reduced task efficiency. Third, the \emph{scalability of multi-agent systems} is hindered by communication bottlenecks, memory overhead, and the exponential growth of action interdependencies as the number of agents increases. Finally, \emph{local model limitations and memory inconsistencies} further degrade system performance in large-scale tasks. For instance, current embodied AI agent systems such as CoELA~\cite{zhang2024building}, COMBO~\cite{zhang2024combo}, and MindAgent~\cite{gong2023mindagent} require 18, 23, and 21 minutes, respectively, even on desktop GPUs, to complete a single long-horizon multi-objective task.

In this work, we present the first comprehensive analysis of latency, efficiency, and scalability of embodied AI systems from \emph{computing and system perspective}. Specifically, we address three key research questions:

\begin{enumerate}
    \item \emph{\textbf{What are the fundamental building blocks and paradigms of generative AI-based embodied systems?}}
    \item \emph{\textbf{What are the system characteristics and sources of inefficiencies in these AI-driven embodied agents?}}
    \item \emph{\textbf{How can the efficiency and scalability of these embodied systems be systematically improved?}}
\end{enumerate}

To address these questions, we start by presenting a systematic categorization of embodied AI systems (Sec.~\ref{sec:paradigm}). We outline the fundamental building blocks---sensing, planning, memory, communication, reflection, and execution---that constitute embodied agents, and categorize them into single-agent and multi-agent systems. 
Single-agent systems adopt either a modularized paradigm, where building blocks are integrated into a pipeline to output an agent's actions, or an end-to-end paradigm, where models directly output actions.
Multi-agent systems can be categorized into centralized and decentralized paradigms, based on how agents cooperate and communicate.

To systematically conduct our analysis and enable future research, we curate a generative AI embodied system workload suite (Sec.~\ref{sec:selected_system}). This workload suite covers several important characteristics of embodied AI systems, including: (1) single-agent and multi-agent architectures, (2) centralized and decentralized coordination paradigms, (3) various task complexities and horizons, (4) different memory capacities and retrieval mechanisms, and (5) diverse communication and planning strategies. The suite provides a robust foundation for in-depth analysis and optimization of embodied AI systems. 

We leverage the workload suite to conduct a study of the characteristics and performance of embodied AI systems across various configurations and scenarios (Sec.~\ref{sec:profiling}). This approach allows us to systematically evaluate the impact of different architectural choices, coordination strategies, and system parameters on overall performance and efficiency (Sec.~\ref{sec:profiling_single_agent} and \ref{sec:profiling_scalability}). Our characterization provides new observations and insights:

\begin{itemize}
\item \textbf{Performance and Latency:} We observed significant runtime latencies for long-horizon multi-objective tasks across the majority of systems, substantially impacting their viability for real-time applications.

\item \textbf{Bottlenecks:} LLM-based planning and communication emerged as the dominant contributors to overall latency, and low-level execution also introduced notable delays. Sequential processing significantly increased task completion times compared to optimized parallel execution.

\item \textbf{Memory vs. Reliability Trade-offs:} Large memory modules generally improve task success rates, but they also introduce increased retrieval latency. We observe memory inconsistencies in a considerable portion of multistep tasks, which affects the overall reliability of the system.

\item \textbf{Scalability Issues:} Multi-agent scenarios reveal distinct challenges: centralized paradigms showed decreased performance as agent count increased, while decentralized paradigms faced escalating communication overhead.

\item \textbf{Component Impacts:} Reflection modules demonstrated improved error correction capabilities with minimal overhead. The execution modules prove critical for the successful completion of tasks in a wide range of scenarios.

\end{itemize}

To the best of our knowledge, this is one of the \emph{first} works to comprehensively assess the efficiency of generative AI-based embodied AI systems from computing and system perspectives. Our detailed analysis and set of recommendations are meant to inspire the design of next-generation embodied AI systems through synergistic advancements in models, software, and computing architectures, ultimately paving the way for more robust and efficient embodied AI architectures.

\begin{figure*}[t!]
\centering\includegraphics[width=2.05\columnwidth]{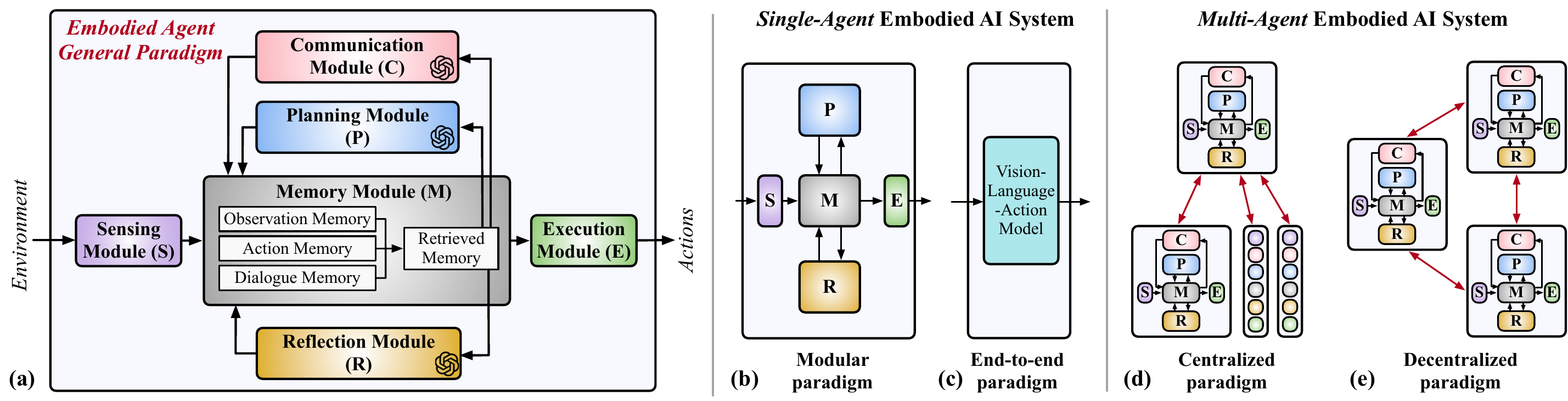}
\vspace{-10pt}
        \caption{\textbf{Embodied AI Agents Paradigm.} \textbf{(a)} The embodied AI agents typically consist of six key modules, where \emph{sensing module} perceives the environment, \emph{planning module} makes high-level plans, \emph{communication module} generates messages, \emph{memory module} stores the agent’s action, dialogue and world knowledge, \emph{execution module} generates primitive actions, and \emph{reflection module} reflects actions. The single-agent embodied AI systems can be built using the \textbf{(b)} modularized-based paradigm or \textbf{(c)} end-to-end pre-trained vision-language-action model. The multi-agent embodied AI systems can be built in the \textbf{(d)} centralized paradigm or \textbf{(e)} decentralized paradigm.}
        \label{fig:paradigm_overview}
        \vspace{-10pt}
\end{figure*}

\section{Embodied AI Agents Paradigm}
\label{sec:paradigm}

This section introduces the computing paradigms of embodied AI agent systems. We first present the general building blocks of embodied agents (Sec.~\ref{subsec:general_paradigm}), then categorize single-agent embodied systems into modularized (Sec.~\ref{subsec:single_agent_modular}) and end-to-end paradigm (Sec.~\ref{subsec:single_agent_e2e}), and categorize multi-agent embodied systems into centralized (Sec.~\ref{subsec:multi_agent_centralized}) and decentralized paradigm (Sec.~\ref{subsec:multi_agent_decentralized}), each accompanied by representative workloads.


\subsection{Building Blocks for Embodied AI Agent System }
\label{subsec:general_paradigm}
Embodied AI agents typically consist of six key modules---sensing, planning, communication, memory, reflection, and execution. 
The \emph{sensing module} is responsible for perceiving the environment and extracting important data. 
The \emph{planning module} decomposes long-horizon task and generates high-level plans and instructions.
The \emph{communication module} manages an agent's information sharing using dialogue generation and comprehension capabilities.
The \emph{execution module} executes the high-level plan by generating low-level primitive actions. 
The \emph{reflection module} reflects erroneous or inefficient actions and prevents hallucinations.
The \emph{memory module} stores the environment observations, agent actions, and dialogue histories during the entire embodied task procedure.
Fig.~\ref{fig:paradigm_overview}\blue{a} presents the building blocks of embodied AI agents.

\textbf{Sensing module.} This module processes sensor data, allowing agents to collect and analyze surrounding information that is critical for higher-level reasoning~\cite{yang2024embodied,zhang2024agent}. It establishes a global or shared environmental model that includes a map of spatial layout, moving entities, obstacles, and resource locations~\cite{min2021film,ramakrishnan2022poni,blukis2022persistent,chen2025octocache}. The agent continuously updates its view of the environment with the task proceeding. In multi-agent scenarios, agents can also communicate their local views with other agents. 

\textbf{Planning module.} The planning module decomposes a long-horizon task into a sequence of sub-objectives.
It begins with retrieving relevant information from the memory module and converting it into text descriptions. It then compiles all potential high-level plans based on the current state and the procedural knowledge stored, allowing the LLMs to make an informed decision~\cite{huang2022language,guan2023leveraging,chen2023robogpt,wang2023describe}. By formalizing the action list, the LLMs can focus on reasoning and generate an executable plan without the need for few-shot demonstrations. To enhance reasoning, prompting techniques (e.g., chain-of-thought~\cite{kojima2022large}, tree-of-thought~\cite{yao2024tree}, and graph-of-though~\cite{besta2024graph}) can be employed, guiding LLMs through a more thorough deliberation process before making a final decision.


\textbf{Communication Module.} The communication module accesses relevant data from memory (e.g., environmental maps, task progress, agent states, and past interactions) and generates appropriate messages for communication with other agents in multi-agent systems~\cite{li2023camel,chen2024scalable,guo2024embodied}. The communication module typically relies on LLM's advanced dialogue generation and comprehension capabilities.

\textbf{Memory module.} Embodied agents store knowledge and experiences from their interactions with the environment and other agents in the memory module, typically categorized into observation, dialogue, and action memory. \emph{Observation memory}~\cite{xiang2024language,song2023llm,gong2023mindagent} holds the agent’s understanding of the world, including maps, task progress, and the states of the agent and others. This memory is updated with new data processed by the sensing module. \emph{Dialogue memory}~\cite{zhang2024building,guo2024embodied} logs the agent’s history of interactions and exchanged dialogue. Each time the agent sends or receives a message from other agents, the relevant information is added to this memory. \emph{Action memory}~\cite{wang2024jarvis,tan2024towards,ahn2022can} records the agents' actions and status, and contains knowledge on how to execute specific high-level plans in different environments, encoded in the form of code or neural model parameters.

\textbf{Reflection module.} Embodied agents may generate unexpected operations and even produce severe hallucinations. The reflection module typically observes the state before and after a decision agent's operation to determine whether the current plan meets expectations~\cite{mandi2024roco,liu2024coherent,zhang2024self}. Based on the reflection results (e.g., erroneous, ineffective, and correct), the agent will choose to re-plan or not and update the correct operation information in the memory module.

\textbf{Execution module.} Although LLMs excel at high-level planning, they are less effective at handling low-level planning and control tasks~\cite{wu2023plan,zhao2024large}. To ensure effective and adaptable decision-making across various environments, the execution module generates primitive actions to carry out the high-level plans in a robust low-level manner~\cite{garrett2021integrated}. This approach allows the planning module to remain generalizable, focusing on broader task-solving while leveraging the LLMs’ extensive world knowledge and reasoning capabilities.

\begin{table}[t!]
\centering
\caption{\textbf{Embodied AI Agent Systems.} Categorization of recent embodied AI agent systems into four paradigms with their computing module compositions. Action types V, T, E represent virtual action, tool usage, and physical action, respectively.} 
\renewcommand*{\arraystretch}{1.15} 
\setlength\tabcolsep{1pt} 
\resizebox{\columnwidth}{!}{%
\begin{tabular}{c|c|c|cccccc|c}
\hline
\multicolumn{2}{c|}{\multirow{2}{*}{\textbf{System Paradigm}}} & \multirow{2}{*}{\textbf{Workloads}} & \multicolumn{6}{c|}{\textbf{Computing Modules}} & \multirow{2}{*} {\textbf{Embodied Type}} \\
\cline{4-9}
\multicolumn{2}{c|}{} & & Sense & Plan & Comm. & Mem. & Refl. & Exec.  \\ \hline

\multirow{25}{*}{\begin{tabular}{c}Single-\\Agent\end{tabular}} 
& \multirow{20}{*}{\begin{tabular}{c}Modularized\\Paradigm\end{tabular}} 
& Mobile-Agent \cite{wang2024mobile} & \green{\cmark} & \green{\cmark} & \color{orange}{\xmark} & \color{orange}{\xmark} & \green{\cmark} & \green{\cmark} & Device Control (T) \\ \cline{3-10}
& & AppAgent \cite{zhang2023appagent} & \green{\cmark} & \green{\cmark} & \color{orange}{\xmark} & \color{orange}{\xmark} & \color{orange}{\xmark} & \green{\cmark} & Device Control (T) \\ \cline{3-10}
& & PDDL \cite{guan2023leveraging} & \color{orange}{\xmark} & \green{\cmark} & \color{orange}{\xmark} & \color{orange}{\xmark} & \green{\cmark} & \color{orange}{\xmark} & Simulation (V) \\ \cline{3-10}
& & RoboGPT \cite{chen2023robogpt} & \green{\cmark} & \green{\cmark} & \color{orange}{\xmark} & \color{orange}{\xmark} & \color{orange}{\xmark} & \green{\cmark} & Simulation (V) \\ \cline{3-10}
& & VOYAGER \cite{wang2023voyager} & \color{orange}{\xmark} & \green{\cmark} & \color{orange}{\xmark} & \green{\cmark} & \green{\cmark} & \green{\cmark} & Simulation (V) \\ \cline{3-10}
& & MP5 \cite{qin2024mp5}& \green{\cmark} & \green{\cmark} & \color{orange}{\xmark} & \color{orange}{\xmark} & \green{\cmark}  & \green{\cmark}  & Simulation (V) \\ \cline{3-10}
& & RILA \cite{yang2024rila} & \green{\cmark} & \green{\cmark} & \color{orange}{\xmark} & \green{\cmark} & \green{\cmark} & \green{\cmark} & Navigation (V) \\ \cline{3-10}
& & CRADLE \cite{tan2024towards} & \green{\cmark} & \green{\cmark} & \color{orange}{\xmark} & \green{\cmark} & \green{\cmark} & \green{\cmark} & Device Control (T) \\ \cline{3-10}
& & STEVE \cite{zhao2025see} & \green{\cmark} & \green{\cmark} & \color{orange}{\xmark} & \color{orange}{\xmark} & \color{orange}{\xmark} & \green{\cmark} & Simulation (V) \\ \cline{3-10}
& & DEPS \cite{wang2023describe} & \green{\cmark} & \green{\cmark} & \color{orange}{\xmark} & \color{orange}{\xmark} & \green{\cmark} & \green{\cmark} & Simulation (V) \\ \cline{3-10}
& & JARVIS-1 \cite{wang2024jarvis} & \green{\cmark} & \green{\cmark} & \color{orange}{\xmark} & \green{\cmark} & \green{\cmark} & \green{\cmark} & Simulation (V) \\ \cline{3-10}
& & FILM \cite{min2021film} & \green{\cmark} & \green{\cmark} & \color{orange}{\xmark} & \color{orange}{\xmark} & \color{orange}{\xmark} & \green{\cmark} & Simulation (V) \\ \cline{3-10}
& & LLM-Planner \cite{song2023llm} & \color{orange}{\xmark} & \green{\cmark} & \color{orange}{\xmark} & \color{orange}{\xmark} & \green{\cmark} & \green{\cmark} & Simulation (V) \\ \cline{3-10}
& & EmbodiedGPT \cite{mu2024embodiedgpt} & \green{\cmark} & \green{\cmark} & \color{orange}{\xmark} & \color{orange}{\xmark} & \color{orange}{\xmark} & \green{\cmark}  & Simulation (V) \\ \cline{3-10}
& & Dadu-E \cite{sun2024dadu} & \green{\cmark} & \green{\cmark} & \color{orange}{\xmark} & \green{\cmark} & \green{\cmark} & \green{\cmark} & Simulation (V) \\ \cline{3-10}
& & MINEDOJO \cite{fan2022minedojo} & \green{\cmark} & \green{\cmark} & \color{orange}{\xmark} & \green{\cmark} & \color{orange}{\xmark} & \green{\cmark}  & Simulation (V) \\ \cline{3-10}
& & Luban \cite{guo2024luban} & \green{\cmark} & \green{\cmark} & \color{orange}{\xmark} & \green{\cmark} & \green{\cmark} & \green{\cmark} & Simulation (V) \\ \cline{3-10}
& & MetaGPT \cite{hong2023metagpt}& \color{orange}{\xmark} & \green{\cmark} & \green{\cmark} & \green{\cmark} & \green{\cmark} & \green{\cmark} & Programming (T) \\ \cline{3-10}
& & Mobile-Agent-V2 \cite{wang2024mobileV2} & \green{\cmark} & \green{\cmark} & \color{orange}{\xmark} & \green{\cmark} & \green{\cmark} & \green{\cmark} & Device Control (T) \\ 
\cline{2-10}


& \multirow{6}{*}{\begin{tabular}{c}End-to-End\\Paradigm\end{tabular}} 
& RT-2 \cite{brohan2023rt} & \multicolumn{6}{c|}{Vision-Language-Action Model} & Robot Control (E) \\ \cline{3-10}
& & RoboVLMs \cite{li2024generalistrobotpoliciesmatters} & \multicolumn{6}{c|}{Vision-Language-Action Model} & Robot Control (E) \\ \cline{3-10}
& & GAIA-1 \cite{hu2023gaia} & \multicolumn{6}{c|}{Generative World Model} & Autonomous Driving (E) \\ \cline{3-10}
& & 3D-VLA \cite{zhen20243d} & \multicolumn{6}{c|}{3D Vision-Language-Action Model} & Robot Control (E) \\ \cline{3-10}
& & Octo \cite{team2024octo} & \multicolumn{6}{c|}{Vision-Language Model + Exec Policy} & Robot Control (E) \\ \cline{3-10}
& & Diffusion Policy \cite{chi2023diffusion} & \multicolumn{6}{c|}{Diffusion Policy} & Robot Control (E) \\
\hline

\multirow{17}{*}{\begin{tabular}{c}Multi-\\Agent\end{tabular}} 
& \multirow{8}{*}{\begin{tabular}{c}Centralized\\Paradigm\end{tabular}} 
& LLaMAC \cite{zhang2023controlling} & \color{orange}{\xmark} & \green{\cmark} & \green{\cmark} & \green{\cmark} & \color{orange}{\xmark} & \green{\cmark} & Simulation (V) \\ \cline{3-10}
& & MindAgent \cite{gong2023mindagent} & \color{orange}{\xmark} & \green{\cmark} & \green{\cmark} & \green{\cmark}  & \color{orange}{\xmark} & \green{\cmark} & Simulation (V) \\ \cline{3-10}
& & OLA \cite{guo2024embodied} & \color{orange}{\xmark} & \green{\cmark} & \green{\cmark} & \green{\cmark} & \green{\cmark} & \green{\cmark} & Simulation (V) \\ \cline{3-10}
& & ALGPT \cite{zhou2024algpt} & \green{\cmark} & \green{\cmark} & \green{\cmark} & \green{\cmark} & \color{orange}{\xmark} & \green{\cmark} & Navigation (V) \\ \cline{3-10}
& & CMAS \cite{chen2024scalable} & \green{\cmark} & \green{\cmark} & \green{\cmark} & \green{\cmark}  & \color{orange}{\xmark} & \green{\cmark} & Simulation (V) \\ \cline{3-10}
& & ReAd \cite{zhang2024towards} & \color{orange}{\xmark} & \green{\cmark} & \green{\cmark} & \color{orange}{\xmark} & \green{\cmark} & \green{\cmark} & Simulation (V) \\ \cline{3-10}
& & Co-NavGPT \cite{yu2023co}& \green{\cmark} & \green{\cmark} & \green{\cmark} & \color{orange}{\xmark} & \color{orange}{\xmark} & \green{\cmark}  & Navigation (V) \\ \cline{3-10}
& & COHERENT \cite{liu2024coherent} & \green{\cmark} & \green{\cmark} & \green{\cmark} & \green{\cmark} & \green{\cmark} & \green{\cmark} & Simulation (V) \\ \cline{2-10}

& \multirow{9}{*}{\begin{tabular}{c}Decentralized\\Paradigm\end{tabular}} 
& DMAS \cite{chen2024scalable} & \green{\cmark} & \green{\cmark} & \green{\cmark} & \green{\cmark} & \color{orange}{\xmark} & \green{\cmark} & Simulation (V) \\ \cline{3-10}
& & HMAS \cite{chen2024scalable} & \green{\cmark} & \green{\cmark} & \green{\cmark} & \green{\cmark} & \green{\cmark} & \green{\cmark} & Simulation (V) \\ \cline{3-10}
& & AGA \cite{yu2024affordable} & \green{\cmark} & \green{\cmark} & \green{\cmark} & \green{\cmark} & \green{\cmark} & \green{\cmark} & Simulation (V) \\ \cline{3-10}
& & CoELA \cite{zhang2024building}& \green{\cmark} & \green{\cmark} & \green{\cmark} & \green{\cmark} & \color{orange}{\xmark} & \green{\cmark} & Simulation (V) \\ \cline{3-10}
& & FMA \cite{lu2023building} & \color{orange}{\xmark} & \green{\cmark} & \green{\cmark} & \green{\cmark} & \green{\cmark} & \green{\cmark} & Programming (T) \\ \cline{3-10}
& & COMBO \cite{zhang2024building}& \green{\cmark} & \green{\cmark} & \green{\cmark} & \green{\cmark} & \color{orange}{\xmark} & \green{\cmark} & Simulation (V) \\ \cline{3-10}
& & RoCo \cite{mandi2024roco}& \green{\cmark} & \green{\cmark} & \green{\cmark} & \green{\cmark} & \green{\cmark} & \green{\cmark} & Simulation (V) \\ \cline{3-10}
& & AgentVerse \cite{chen2023agentverse} & \color{orange}{\xmark} & \green{\cmark} & \green{\cmark} & \color{orange}{\xmark} & \color{orange}{\xmark} & \green{\cmark} & Simulation (V) \\ \cline{3-10}
& & KoMA \cite{jiang2024koma} & \color{orange}{\xmark} & \green{\cmark} & \green{\cmark} & \green{\cmark} & \green{\cmark} & \green{\cmark} & Simulation (V) \\
\hline   
\end{tabular}
}
\vspace{-10pt}
\end{table}

\subsection{Single-Agent Embodied AI System: Modularized Paradigm}
\label{subsec:single_agent_modular}
A number of embodied AI systems leverage the modularized paradigm to build agents (Fig.~\ref{fig:paradigm_overview}\blue{b}). The single-agent systems are usually built upon a partial of building blocks (e.g., sensing, planning, memory, reflection, and execution) as detailed in Sec.~\ref{subsec:general_paradigm} to conduct long-horizon multi-objective task and motion planning. For instance, 
STEVE~\cite{zhao2025see}, AppAgent~\cite{zhang2023appagent} and RoboGPT~\cite{chen2023robogpt} consist of three building blocks (sensing, planning, execution); 
DEPS~\cite{wang2023describe}, MP5~\cite{qin2024mp5} and Mobile-Agent~\cite{wang2024mobile} consist of four building blocks (sensing, planning, reflection, execution);
MINEDOJO~\cite{fan2022minedojo} consists of four building blocks (sensing, planning, memory, execution); 
CRADLE~\cite{tan2024towards}, RILA~\cite{yang2024rila}, JARVIS-1~\cite{wang2024jarvis}, and Dadu-E~\cite{sun2024dadu} consist of five building blocks (sensing, planning, memory, reflection, execution). In addition to these workloads, MetaGPT~\cite{hong2023metagpt} and Mobile-Agent-V2~\cite{wang2024mobileV2} may also be characterized as single-agent systems encompassing multiple agent roles.

\subsection{Single-Agent Embodied AI System: End-to-End Paradigm}
\label{subsec:single_agent_e2e}
Besides the modularized paradigm, the end-to-end paradigm that directly outputs agent actions from the models is another approach for short-horizon embodied tasks ~\cite{brohan2023rt,li2024generalistrobotpoliciesmatters, zhen20243d, team2024octo, chi2023diffusion}. The end-to-end approach typically relies on general-purpose visually-aligned large language models trained on large-scale text, image, and video data to serve as a foundation for creating embodied multi-modal agents that can act in various environments (Fig.~\ref{fig:paradigm_overview}\blue{c}). For instance, RT-2~\cite{brohan2023rt} is a vision-language-action model pre-trained on web-scale knowledge to facilitate end-to-end robotic control. RoboVLMs~\cite{li2024generalistrobotpoliciesmatters} introduces a framework designed to transform vision-language models into versatile vision-language-action models. 3D-VLA~\cite{zhen20243d} extends this paradigm by incorporating 3D inputs, enabling seamless integration with the broader 3D physical world. GAIA-1~\cite{hu2023gaia} leverages video, text, and action inputs to generate realistic driving scenarios with fine-grained control over ego-vehicle behavior and scene features.

\subsection{Multi-Agent Embodied AI System: Centralized Paradigm}
\label{subsec:multi_agent_centralized}
Based on the inspiring capabilities of the single embodied agent, multi-agent embodied AI systems have been developed to leverage collective intelligence and specialized profiles and skills of multiple agents. In multi-agent systems, each agent typically consists of all or part of the building blocks (sensing, planning, communication, memory, reflection, and execution) to collaboratively engage in planning, discussions, and decision-making, mirroring the cooperative nature of human group work in long-horizon multi-objective tasks.

One option to build a multi-agent embodied system is where a centralized agent generates and communicates the next step plan for all embodied agents in the system, and each agent can provide local feedback to the central agent planner (Fig.~\ref{fig:paradigm_overview}\blue{d}). For instance, MindAgent~\cite{gong2023mindagent}, OLA~\cite{guo2024embodied}, CMAS~\cite{chen2024scalable}, and COHERENT~\cite{liu2024coherent} leverage centralized paradigm to collaboratively conduct gaming interaction and long-horizon task and motion planning.

\subsection{Multi-Agent Embodied AI System: Decentralized Paradigm}
\label{subsec:multi_agent_decentralized}
The multi-agent embodied system can also be built in a decentralized paradigm in which each agent generates its plan and engages in collaborative dialogue with other agents (Fig.~\ref{fig:paradigm_overview}\blue{e}). Similarly, each agent is equipped with sensing, planning, communication, memory, reflection, and execution modules. For instance, CoELA~\cite{zhang2024building}, COMBO~\cite{zhang2024combo}, DMAS~\cite{chen2024scalable}, RoCo~\cite{mandi2024roco}, Organized LLM Agents~\cite{guo2024embodied}, and KoMA~\cite{jiang2024koma} leverages decentralized paradigm to accomplish long-horizon tasks such object transport, robotic manipulation, autonomous vehicle control, and gaming interaction.

\section{Embodied Agent Systems Workload Suite}
\label{sec:selected_system}

This section presents our workload suite for benchmarking embodied agent systems (Sec.~\ref{subsec:benchmark_overview}, \ref{subsec:selected_single_modular}, \ref{subsec:benchmark_centralized}, \ref{subsec:benchmark_decentralized}), with the hardware setup and key benchmark metrics (Sec.~\ref{subsec:profiling_setup}).


\begin{figure*}
\vspace{-6pt}
\begin{minipage}[b]{\linewidth}
    \centering
    \includegraphics[width=\columnwidth]{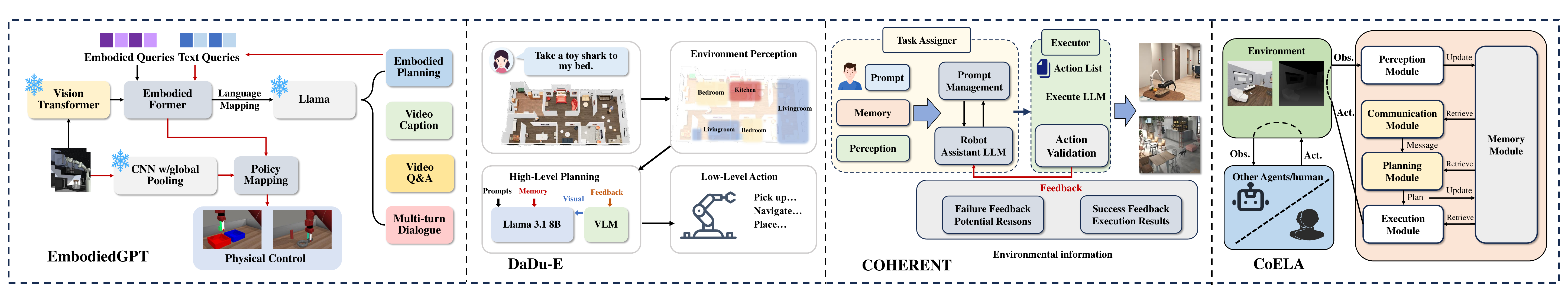}
\end{minipage}%
 \vspace{-1pt}
\begin{minipage}[b]{\linewidth}
    \renewcommand*{\arraystretch}{1.2}
    \setlength\tabcolsep{2.5pt}
\resizebox{\linewidth}{!}{%
\centering
\begin{threeparttable}
\begin{tabular}{c|cccccc|c|c|c|c}
\hline
\multirow{2}{*}{\textbf{Embodied AI Systems}} & \multicolumn{6}{c|}{\textbf{System Module}}                                      & \multirow{2}{*}{\textbf{Application}} & \multirow{2}{*}{\textbf{Datasets and Tasks}} & \multirow{2}{*}{\textbf{\begin{tabular}[c]{@{}c@{}}Single/Multi-\\ Agent\end{tabular}}} & \multirow{2}{*}{\textbf{\begin{tabular}[c]{@{}c@{}}Centralized/\\ Decentralized\end{tabular}}} \\ \cline{2-7}
                                              & \multicolumn{1}{c|}{Sensing} & \multicolumn{1}{c|}{Planning} & \multicolumn{1}{c|}{Communication} & \multicolumn{1}{c|}{Memory\tnote{*}} & \multicolumn{1}{c|}{Reflection} & \multicolumn{1}{c|}{Execution} &                                       &                                                                                         &                                                                                                \\ \hline
EmbodiedGPT~\cite{mu2024embodiedgpt}  & \multicolumn{1}{c|}{ViT} & \multicolumn{1}{c|}{Llama-7B}&  \multicolumn{1}{c|}{--}&  \multicolumn{1}{c|}{--}&  \multicolumn{1}{c|}{--} & MLP & Embodied planning, visual captioning, VQA & Franka Kitchen~\cite{gupta2019relay}, Meta-World~\cite{yu2020meta}, VirtualHome~\cite{puig2018virtualhome} & Single-Agent & -- \\ \hline

JARVIS-1~\cite{wang2024jarvis} & \multicolumn{1}{c|}{MineCLIP} & \multicolumn{1}{c|}{GPT-4/Llama-13B} & \multicolumn{1}{c|}{--} & \multicolumn{1}{c|}{Ob., Act.} & \multicolumn{1}{c|}{Llama-13B} & Action list & Embodied planning (e.g, obtain diamond pickaxe)  & Minecraft~\cite{lin2023mcu} & Single-Agent & -- \\ \hline

DaDu-E~\cite{sun2024dadu}  & \multicolumn{1}{c|}{PointCloud} & \multicolumn{1}{c|}{Llama-8B} & \multicolumn{1}{c|}{--} & \multicolumn{1}{c|}{Ob., Act.} & \multicolumn{1}{c|}{LLaVA-8B} & AnyGrasp & Object transport, Autonomous decision-making & Self-designed four-level tasks & Single-Agent & -- \\ \hline

MP5~\cite{qin2024mp5}  & \multicolumn{1}{c|}{MineCLIP} & \multicolumn{1}{c|}{GPT-4} &  \multicolumn{1}{c|}{--} & \multicolumn{1}{c|}{--}  & \multicolumn{1}{c|}{GPT-4} & \multicolumn{1}{c|}{MineDojo} & Object transport, Situation-aware long-term planning & Minecraft~\cite{lin2023mcu} & Single-Agent & -- \\ \hline

DEPS~\cite{wang2023describe}  &  \multicolumn{1}{c|}{Symbolic info} & \multicolumn{1}{c|}{GPT-4} & \multicolumn{1}{c|}{--} & \multicolumn{1}{c|}{--} & \multicolumn{1}{c|}{CLIP} & MineDojo & Embodied planning (e.g, obtain diamond pickaxe) & Minecraft~\cite{lin2023mcu}, MineRL~\cite{baker2022video}, ALFWorld~\cite{shridhar2020alfworld} & Single-Agent & --\\ \hline


MindAgent~\cite{gong2023mindagent}  & \multicolumn{1}{c|}{--}  & \multicolumn{1}{c|}{GPT-4} & \multicolumn{1}{c|}{GPT-4}  & \multicolumn{1}{c|}{Ob., Act., Dx.} & \multicolumn{1}{c|}{--} & Action list & Collaborative planning, gaming, housework& CuisineWorld~\cite{gong2023mindagent}, Minecraft~\cite{lin2023mcu} & Multi-Agent & Centralized \\ \hline

OLA~\cite{guo2024embodied}  & \multicolumn{1}{c|}{--} & \multicolumn{1}{c|}{GPT-4/Llama-70B} & \multicolumn{1}{c|}{GPT-4} & \multicolumn{1}{c|}{Ob., Act., Dx.} & \multicolumn{1}{c|}{GPT-4} & \multicolumn{1}{c|}{Action list} & Collaborative planning, object transport & VirtualHome~\cite{puig2018virtualhome}, C-WAH~\cite{puig2021watch} & Multi-Agent & Centralized \\ \hline

COHERENT~\cite{liu2024coherent}  & \multicolumn{1}{c|}{DINO} & \multicolumn{1}{c|}{GPT-4} & \multicolumn{1}{c|}{GPT-4} & \multicolumn{1}{c|}{Ob., Act., Dx.} & \multicolumn{1}{c|}{GPT-4} & RRT/A-star & Collaborative planning, Robot arm manipulation & BEHAVIOR-1K~\cite{li2023behavior} & Multi-Agent & Centralized \\ \hline

CMAS~\cite{chen2024scalable}  & \multicolumn{1}{c|}{ViLD} & \multicolumn{1}{c|}{GPT-4} & \multicolumn{1}{c|}{GPT-4} & \multicolumn{1}{c|}{Ob., Act., Dx.} & \multicolumn{1}{c|}{--} & Action list & Collaborative planning, manipulator, object transport & BoxNet1, BoxNet2, WareHouse, BoxLift~\cite{chen2024scalable} & Multi-Agent & Centralized \\ \hline

CoELA~\cite{zhang2024building}  & \multicolumn{1}{c|}{Mask R-CNN} & \multicolumn{1}{c|}{GPT-4} & \multicolumn{1}{c|}{GPT-4} & \multicolumn{1}{c|}{Ob., Act., Dx.} & \multicolumn{1}{c|}{--} & \multicolumn{1}{c|}{A-star} & Collaborative object transporting, housework & TDW-MAT~\cite{gan2022threedworld}, C-WAH~\cite{puig2021watch} & Multi-Agent & Decentralized \\ \hline

COMBO~\cite{zhang2024combo}  & \multicolumn{1}{c|}{Diffusion}  & \multicolumn{1}{c|}{LLaVA-7B} & \multicolumn{1}{c|}{LLaVA-7B} & \multicolumn{1}{c|}{Ob., Act., Dx.} & \multicolumn{1}{c|}{--} & \multicolumn{1}{c|}{A-star} & Collaborative gaming, housework & TDW-Game~\cite{gan2021threedworld}, TDW-Cook~\cite{gan2021threedworld} & Multi-Agent & Decentralized \\ \hline

RoCo~\cite{mandi2024roco}  & \multicolumn{1}{c|}{ViT} & \multicolumn{1}{c|}{GPT-4} & \multicolumn{1}{c|}{GPT-4} & \multicolumn{1}{c|}{Ob., Act., Dx.} & \multicolumn{1}{c|}{GPT-4} & RRT & Robot arm motion planning, manipulation & RoCoBench~\cite{mandi2024roco} & Multi-Agent & Decentralized \\ \hline

DMAS~\cite{chen2024scalable}  & \multicolumn{1}{c|}{ViLD} & \multicolumn{1}{c|}{GPT-4} & \multicolumn{1}{c|}{GPT-4} & \multicolumn{1}{c|}{Ob., Act., Dx.} & \multicolumn{1}{c|}{--} & Action list & Collaborative planning, manipulator, object transport & BoxNet1, BoxNet2, WareHouse, BoxLift~\cite{chen2024scalable} & Multi-Agent & Decentralized \\ \hline

HMAS~\cite{chen2024scalable}  & \multicolumn{1}{c|}{ViLD} & \multicolumn{1}{c|}{GPT-4} & \multicolumn{1}{c|}{GPT-4} & \multicolumn{1}{c|}{Ob., Act., Dx.} & \multicolumn{1}{c|}{GPT-4} & Action list & Collaborative planning, manipulator, object transport & BoxNet1, BoxNet2, WareHouse, BoxLift~\cite{chen2024scalable} & Multi-Agent & Decentralized \\ \hline

\end{tabular}
\begin{tablenotes}
        \item[*] Ob.: observation memory; Act.: Action memory; Dx.: dialog memory. \vspace{-10pt}
      \end{tablenotes}
    \end{threeparttable}
}
\captionof{table}{\textbf{Embodied Agent Systems Workload Suite.} Above are the workflow examples of four embodied AI agent systems. Below are the details of the benchmarked embodied agent systems, including the models used for each key building block (perception, planning, communication, memory, reflection, and execution), applications, datasets and tasks, and the extent of single or multi-agent collaboration.}
\vspace{-5pt}
\label{tab:selected_workload}
\end{minipage}
\end{figure*}

\subsection{Workload Suite Overview}
\label{subsec:benchmark_overview}

Our workload suite consists of 14 embodied AI agent systems. As listed in Tab.~\ref{tab:selected_workload}, each embodied system targets a specific usecase and specifies building block modules, applications, deployment scenarios, and paradigms. Additionally, we implement each system as a docker image for better portability. 
The 14 embodied AI agent systems include five single-agent systems (EmbodiedGPT~\cite{mu2024embodiedgpt}, JARVIS-1~\cite{wang2024jarvis}, DaDu-E~\cite{sun2024dadu}, MP5~\cite{qin2024mp5}, DEPS~\cite{wang2023describe}), four centralized multi-agent systems (MindAgent~\cite{gong2023mindagent}, OLA~\cite{guo2024embodied}, COHERENT~\cite{liu2024coherent}, CMAS~\cite{chen2024scalable}), and five decentralized multi-agent systems (CoELA~\cite{zhang2024building}, COMBO~\cite{zhang2024combo}, RoCo~\cite{mandi2024roco}, DMAS~\cite{chen2024scalable}, HMAS~\cite{chen2024scalable}). These systems represent a variety of embodied paradigms and exhibit state-of-the-art performance on long-horizon tasks. Interested readers could refer to their references for system details.

\subsection{Single-Agent Embodied Systems}
\label{subsec:selected_single_modular}

For single-agent systems, our analysis mainly focuses on the modularized paradigm since these agents are designed for long-horizon embodied planning tasks.

\textbf{EmbodiedGPT.} \emph{EmbodiedGPT}~\cite{mu2024embodiedgpt} is a multi-modal modularized embodied system for long-horizon embodied tasks. 
It consists of a sensing model (Vision Transformer), a visual-language planning model (fine-tuned Llama-7B model), and a low-level execution policy network (multi-layer perception) that enables seamless integration of high-level planning and low-level control. 
Evaluated on Franka Kitchen~\cite{gupta2019relay}, Meta-World~\cite{yu2020meta}, and Virtual-Home~\cite{puig2018virtualhome}, EmbodiedGPT excels in embodied tasks such as embodied planning, embodied control, visual captioning, and visual question answering.

\textbf{JARVIS-1.} 
\emph{JARVIS-1}~\cite{wang2024jarvis} is an open-world agent that can perceive multimodal input (visual observations and human instructions), generate sophisticated plans, and perform embodied control.
It consists of a sensing model (MineCLIP~\cite{fan2022minedojo}), long-horizon planning (GPT-4 or Llama-13B), a memory module for storing observations and actions, a self-reflection module, and an execution module.
Evaluated on Minecraft~\cite{lin2023mcu}, JARVIS-1 excels in a range of embodied tasks from short-horizon planning (e.g., chopping trees) to long-horizon tasks (e.g., obtaining diamond pickaxe).

\textbf{DaDu-E.}
\emph{DaDu-E}~\cite{sun2024dadu} is a robust closed-loop planning framework for embodied AI robots. 
It is equipped with a sensing module based on LiDAR point cloud, a lightweight planning module (Llama-8B), a reflection module (LLaVA-8B), a memory augmentation, and a low-level grasping execution module (AnyGrasp~\cite{fang2023anygrasp}).
Evaluated on self-designed four-level embodied AI tasks, DaDu-E performs well in multi-task execution, long-horizon decision-making, cognition language comprehension, and object transport.

\textbf{MP5.}
\emph{MP5}~\cite{qin2024mp5} is an open-ended multimodal embodied system that can decompose feasible sub-objectives, design sophisticated situation-aware plans, and perform embodied action control.
It consists of a sensing module (MineCLIP~\cite{fan2022minedojo}), a planning module (GPT-4), a reflection patroller (GPT-4), and a low-level performer (MineDojo~\cite{fan2022minedojo}).
Evaluated on Minecraft~\cite{lin2023mcu}, MP5 performs well on open-end tasks both with heavy process dependency and context dependency.

\textbf{DEPS.}
\emph{DEPS}~\cite{wang2023describe} is a multi-task embodied agent system for solving complex and long-horizon tasks in open-world environments.
It is equipped with a sensing module (symbolic information), a planning module (GPT-4), a reflection module (CLIP), and a low-level controller (MineDojo~\cite{fan2022minedojo}).
Evaluated on Minecraft~\cite{lin2023mcu}, MineRL~\cite{baker2022video}, and ALFWorld~\cite{shridhar2020alfworld}, DEPS exhibit capabilities to deal with complicated tasks with complex dependency and relation in the open-ended world.



\subsection{Multi-Agent Centralized Embodied Systems}
\label{subsec:benchmark_centralized}

\textbf{MindAgent.} 
\emph{MindAgent}~\cite{gong2023mindagent} is a multi-agent embodied system for collaborative gaming and household tasks, enabling agents to cooperate on complex long-horizon tasks with emergent planning capabilities in a centralized manner. 
LLMs (GPT-4) facilitate task scheduling and cooperation, improving planning efficiency through few-shot prompting and feedback. 
Evaluated on CuisineWorld~\cite{gong2023mindagent} and Minecraft~\cite{lin2023mcu}, MindAgent shows strong performance in enhancing multi-agent coordination and collaboration efficiency.

\textbf{OLA.} 
\emph{Organized LLM Agents (OLA)}~\cite{guo2024embodied} is a multi-agent framework that offers the flexibility to prompt and organize embodied agents into various team structures, facilitating versatile inter-agent communication.
Each agent is equipped with planning and communication modules (GPT-4), conducts criticize-reflect on team performance, and generates improved organizational prompts. 
Evaluated on VirtualHome~\cite{puig2018virtualhome} and C-WAH~\cite{puig2021watch}, OLA is effective in long-horizon planning tasks with reduced communication cost and improved efficiency.

\textbf{COHERENT.}
\emph{Collaboration of Heterogeneous Multi-Robot System (COHERENT)}~\cite{liu2024coherent} is a centralized hierarchical framework for heterogeneous multi-robot task planning.
It consists of a sensing module (DINO~\cite{liu2025grounding}) and proposal-execution-feedback-adjustment mechanism (GPT-4) to decompose the complex task into subtasks, and then assigns subtasks to robot executors (RRT or A-star).
Evaluated across 100 BEHAVIOR-1K scenarios~\cite{li2023behavior}, COHERENT  efficiently accomplishes complex and long-horizon task and motion planning.

\textbf{CMAS.}
\emph{CMAS}~\cite{chen2024scalable} is a centralized multi-agent embodied system for collaborative planning.
It uses the image-to-text model ViLD~\cite{gu2021open} to provide text descriptions for environment objects. A central agent uses GPT-4 to produce the next action for all robots and communicates the instructions.
Evaluated across BoxNet, Warehouse, and BoxLift environments, CMAS excels in long-horizon heterogeneous multi-robot planning.

\subsection{Multi-Agent Decentralized Embodied Systems}
\label{subsec:benchmark_decentralized}

\textbf{CoELA.} \emph{Cooperative Embodied Language Agent (CoELA)}\cite{zhang2024building} is designed to enable embodied agents to collaborate with each other or with humans in decentralized environments for long-horizon multi-objective tasks. 
With Mask R-CNN for perception, GPT-4 for planning and communication, CoELA demonstrates strong performance on TDW-MAT~\cite{gan2022threedworld} and C-WAH~\cite{puig2021watch} tasks, excelling at perceiving complex environments, reasoning about world and other agents, communicating efficiently, and executing long-horizon plans in tasks such as collaborative object transport and housework.

\textbf{COMBO.} \emph{Compositional Model for Embodied Multi-Agent Cooperation (COMBO)}~\cite{zhang2024combo} is an embodied multi-agent decentralized system based on compositional world models to facilitate online cooperative planning. Agents reconstruct the global world state from partial egocentric observations (via diffusion model~\cite{ko2023learning}), and leverage Visual Language Models (LLaVA-1.5~\cite{liu2024visual}) to infer other agents' intents, communicate, and propose actions. 
With refined action sequence through tree search, COMBO demonstrates strong performance in long-horizon cooperation tasks like cooking and puzzle-solving in ThreeDWorld~\cite{gan2021threedworld} (TDW-Game and TDW-Cook).

\textbf{RoCo.}
\emph{RoCo}~\cite{mandi2024roco} is a zero-shot multi-robot embodied system for collaborative manipulation and trajectory planning tasks.
It is equipped with a sensing module (OWL-ViT~\cite{minderer2022simple}), a planning module (GPT-4), a communication module (GPT-4), a memory module, a reflection module (GPT-4), and low-level RRT planner. 
Evaluated on RoCoBench~\cite{mandi2024roco}, RoCo is flexible in handling a large variety of tasks with improved task-level coordination and action-level motion planning.

\textbf{DMAS.}
\emph{DMAS}~\cite{chen2024scalable} is a decentralized multi-agent embodied system for collaborative planning.
Each agent uses GPT-4 for planning and dialogue proceeds in rounds of turn-taking among agents.
Evaluated across BoxNet, Warehouse, and BoxLift environments, DMAS efficiently accomplishes collaborative planning, manipulator, and object transport.

\textbf{HMAS.}
\emph{HMAS}~\cite{chen2024scalable} is a multi-agent system that combines the centralized and decentralized approach where an agent provides an initial plan to prime dialogue between agents, and each agent will provide local feedback to the central agent during task execution.
HMAS exhibits excellent performance in collaborative long-horizon tasks and motion planning.

\subsection{Benchmark Metrics and Hardware Setup}
\label{subsec:profiling_setup}

\textbf{Benchmark metrics.} We evaluate both task performance and system efficiency of each selected embodied AI workload. 
Task performance is measured by the \emph{success rate} metric.
System efficiency is measured by the \emph{runtime latency} and \emph{average steps} metrics, representing the latency per step and the average number of steps taken per task.

\textbf{Hardware setup.} To evaluate the performance of embodied AI workloads, we follow the configuration reported in their respective studies. Each LLM agent is instantiated with the GPT-4 from the OpenAI API, while local model inferences (e.g., LLaVA, Llama, DINO, etc) are executed on NVIDIA A6000 GPU. Agent action is executed on Intel i7 CPU.
\section{Latency and Module Sensitivity Analysis}
\label{sec:profiling}

This section presents the benchmarking analysis of the building blocks in embodied agent systems, from their latency runtime (Sec.~\ref{subsec:profiling_runtime}) and sensitivity characteristics (Sec.~\ref{subsec:profiling_sensitivity}). 

\subsection{Runtime Latency Analysis}
\label{subsec:profiling_runtime}

\begin{figure}[t!]
\centering\includegraphics[width=\columnwidth]{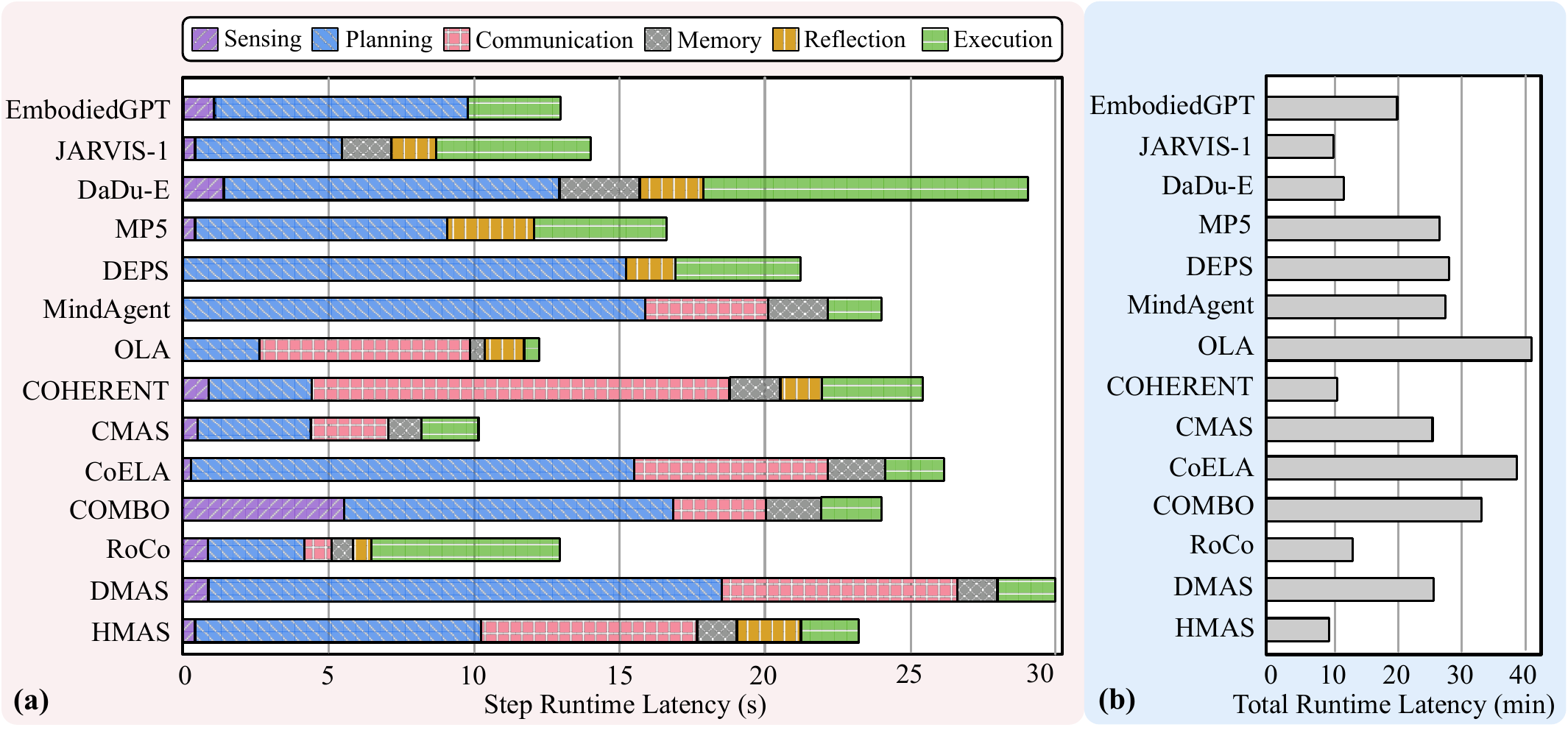}
\vspace{-10pt}
        \caption{\textbf{Runtime Latency Analysis.} \textbf{(a)} Average runtime percentage contributed by each module per time step, and \textbf{(b)} Total runtime latency per task, benchmarked across 14 embodied AI workloads.}
        \label{fig:profiling_latency_breakdown}
        \vspace{-5pt}
\end{figure}

\textbf{End-to-end latency.} Fig.~\ref{fig:profiling_latency_breakdown}\blue{a} illustrates the average latency breakdown contributed by each module per time step in various long-horizon embodied tasks. We observe that \emph{embodied workloads exhibit significant latency step with a low frame rate.} Executing one step of long-horizon tasks takes an average of 10-30 seconds across workloads, leading to a slow frame rate that often fails to meet the real-time requirements of human-agent applications. 
Moreover, as illustrated in Fig.~\ref{fig:profiling_latency_breakdown}\blue{b}, each workload takes tens to hundreds of steps to accomplish the tasks, resulting in \emph{long end-to-end long-horizon embodied task latency}, ranging from 10-40~mins across tasks.

\textbf{LLM-based module latency.} \emph{LLM-based planning and communication modules contribute the most to latency.} As shown in Fig.~\ref{fig:profiling_latency_breakdown}\blue{a}, LLM-based modules, whether through GPT-4 API call or local model processing (e.g., Llama, LLaVA), account for an average of 70.2\% of the total latency across 14 workloads. The communication module significantly bottlenecks certain workloads (e.g., COHERENT, CoELA) due to the repeated processes of message generation and extraction. Additionally, \emph{each execution step often involves multiple LLM inference runs.} For example, in CoELA, each step involves three LLM runs for message generation (16.1\%), planning (36.5\%), and action selection (10.3\%), causing inefficiencies, especially with more agents and longer-horizon tasks.

\textbf{Non-LLM-based module latency.} \emph{Execution modules can become bottlenecks in embodied systems.} Fig.~\ref{fig:profiling_latency_breakdown}\blue{a} reveals that compared to LLM-based modules, the low-level planning and action module latency is not negligible, accounting for 49.4\%, 38.1\%, and 24.1\% of the total latency in RoCo, DaDu-E, and EmbodiedGPT. This is primarily due to the multiple executions typically required to complete a single planned step, as well as computational complexity of low-level path planning and manipulation functions (e.g., RRT~\cite{noreen2016optimal}, A-star~\cite{sanchez2021path}). It is worth noting that low-level execution is essential for the successful completion of embodied tasks~\cite{wu2023plan,zhao2024large}.

\colorbox{lightgray!50}{\textbf{Takeaway 1:}} \textit{
End-to-end latency in long-horizon embodied tasks is significant. LLM-based planning and communication dominate the latency due to repeated inference runs. Low-level planning and execution also contribute notable delays from multiple executions and computational complexity.}

\colorbox{brown!30}{\textbf{Recommendation 1:}} \textit{The long latency of planning and communication can be optimized through efficient LLM deployment, such as batching (e.g., aggregate multiple queries into single batch), quantization (e.g., AWQ~\cite{lin2024awq}), hardware-friendly formats (e.g., MLC-LLM~\cite{mlc-llm}), lightweight models.}

\colorbox{brown!30}{\textbf{Recommendation 2:}} \textit{The inefficiency of low-level action and execution can be optimized via optimized data structure, memory access pattern, parallelism, and domain-specific architecture integrated with high-level planning substrate.}

\begin{figure}[t!]
\centering\includegraphics[width=\columnwidth]{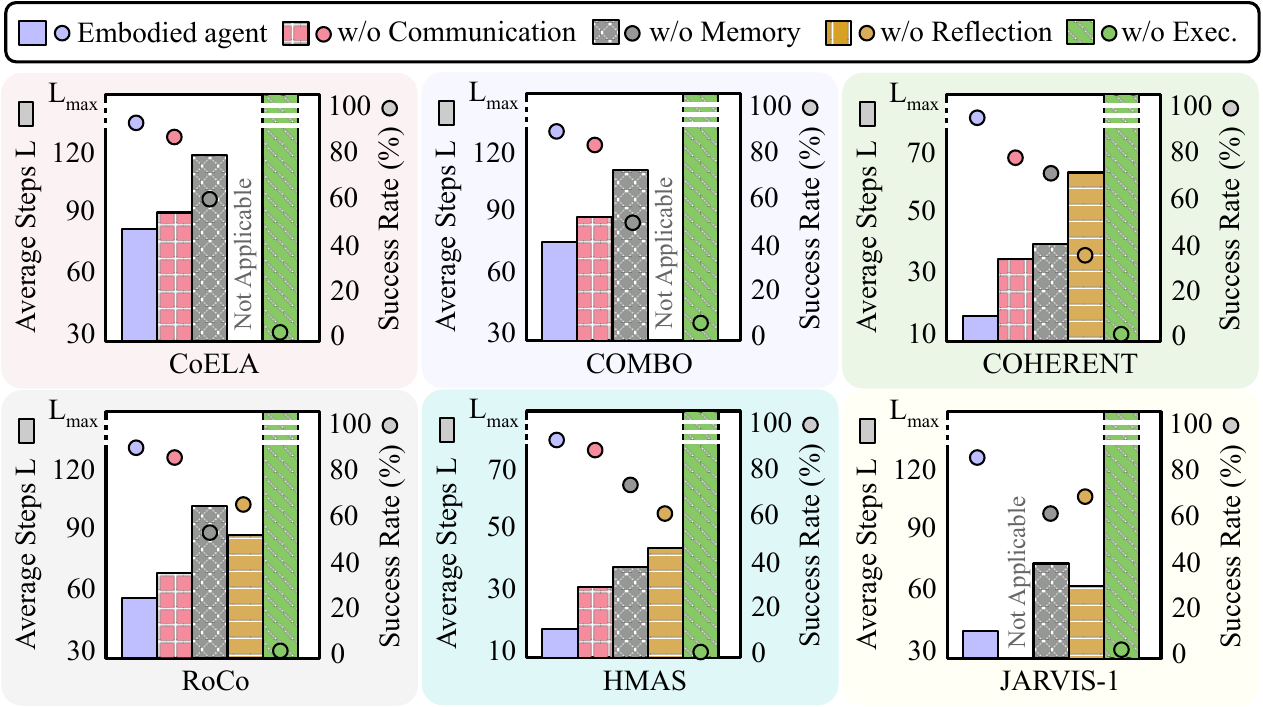}
\vspace{-10pt}
        \caption{\textbf{Module Sensitivity Analysis.} Average success rate and the number of steps taken to complete the long-horizon tasks across six single-agent and multi-agent embodied AI systems.}
        \label{fig:profiling_ablation}
        \vspace{-5pt}
\end{figure}

\subsection{Module Sensitivity Analysis}
\label{subsec:profiling_sensitivity}
To analyze the sensitivity of each building block in embodied systems for long-horizon multi-objective tasks, we benchmark the average number of steps required to complete tasks and the average success rate across workloads (Fig.~\ref{fig:profiling_ablation}).

\textbf{Communication module sensitivity.} The communication module facilitates information sharing and request handling. Interestingly, we observe that disabling communication among agents does not have significant impact on performance compared to other modules. We hypothesize two possible reasons: \emph{At the system level}, many dialogue rounds are redundant and unproductive, with only a small portion of generated messages being exchanged, indicating sparse use of communication. \emph{At the model level}, effective communication relies on accurately modeling other agents’ intentions and resolving natural language ambiguities. Current models (e.g., GPT-4) still struggle to manage these reasoning complexities consistently.

\textbf{Memory module sensitivity.} 
The memory module stores observation, dialogue, and action information, which other modules retrieve for planning and communication. As shown in Fig.~\ref{fig:profiling_ablation}, the memory module plays a critical role in embodied AI agent systems. Disabling it increases the steps required to complete tasks by an average of 1.61$\times$ and reduces the success rate by 27.7\% across six systems. This highlights the memory module's importance in storing and updating knowledge about the environment and agent actions, significantly enhancing task efficiency and performance.

\textbf{Reflection module sensitivity.} The reflection module is essential for correcting erroneous operations, preventing agents from executing incorrect plans or getting stuck in loops of invalid operations. As shown in Fig.~\ref{fig:profiling_ablation}, disabling the reflection module results in 1.88$\times$ increase in the average number of steps and 33.3\% drop in success rate across six systems. This underscores the importance of reflection mechanism, despite it accounting for only 8.61\% of total latency on average.

\textbf{Execution module sensitivity.} The low-level execution and control module is indispensable for system functionality, as shown in Fig.~\ref{fig:profiling_ablation}. Disabling it led to task failures and reaching the maximum step limit. This likely occurred because,  without this module, the LLM-based planning system was forced to handle low-level control decisions, vastly expanding the decision space and slowing down the inference process. This finding underscores the necessity of using LLMs for high-level planning while relying on low-level control for precise robot agent action. Developing agents that can manage low-level controls efficiently is critical for complex embodied systems.

\colorbox{lightgray!50}{\textbf{Takeaway 2:}} \textit{The memory and reflection modules are crucial for task efficiency, and the execution module is essential for low-level control to prevent task failures. The communication module has no significant impact on task success rate due to redundant dialogues and limited effective use.}

\colorbox{brown!30}{\textbf{Recommendation 3:}} \textit{System performance can be optimized by improving communication efficiency, enhancing memory through context summarization, and strengthening reflection with adaptive error correction. Offloading low-level execution to specialized controllers and adopting a hybrid planning framework can further boost task efficiency.}

\section{System Configuration Impact Analysis}
\label{sec:profiling_single_agent}

Building on the characteristics of the building blocks, this section provides an in-depth benchmarking analysis of the impact of architectural and system configurations, including various planning LLM sizes (Sec.~\ref{subsec:profiling_local_api}), memory capacities (Sec.~\ref{subsec:profiling_memory}), communication token lengths (Sec.~\ref{subsec:profiling_prompt_length}), and the embodied system execution pipeline (Sec.~\ref{subsec:profiling_system_inefficiency}).

\begin{figure}[t!]
\centering\includegraphics[width=\columnwidth]{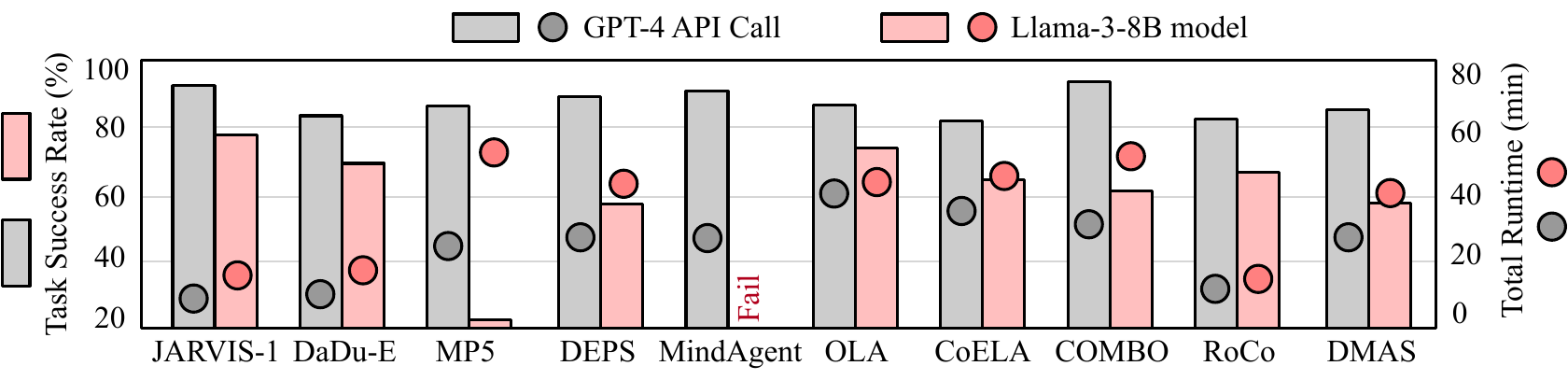}
\vspace{-10pt}
        \caption{\textbf{Local Model Analysis.} Task success rate and end-to-end task runtime under GPT-4 API call and Llama-3-8B local processing.}
        \label{fig:profiling_local_model}
        \vspace{-10pt}
\end{figure}

\subsection{Planning Module: Various LLM Models Analysis}
\label{subsec:profiling_local_api}
\textbf{Local model processing comparison.}
Fig.~\ref{fig:profiling_local_model} compares task performance and system efficiency between local LLM processing (Llama-3.1-8B) and GPT-4 API calls. We observe that \emph{smaller LLM models reduce task success rates and increase end-to-end runtime latency.} This is primarily due to the lower reasoning capability of smaller LLM models, which leads to suboptimal plans, incorrect actions, or requiring more steps to complete tasks. While local models generally reduce per-inference time, the degraded performance and additional actions ultimately result in longer overall task runtime latency.

\colorbox{lightgray!50}{\textbf{Takeaway 3:}} \textit{Smaller local LLMs increase end-to-end embodied task runtime latency and degrade success rates due to suboptimal plans, despite faster per-inference times.}


\colorbox{brown!30}{\textbf{Recommendation 4:}} \textit{Parameter-efficient fine-tuning with LoRA and augmenting smaller LLMs with external knowledge (e.g., symbolic reasoning) can enhance performance and improve task-specific reasoning. Furthermore, converting planning tasks into multiple-choice questions can greatly reduce the complexity of generating format-compliant outputs, thereby narrowing the performance gap between smaller, locally deployed models and closed-source commercial models.}

\subsection{Memory Module: Various Memory Capacities Analysis}
\label{subsec:profiling_memory}

\begin{figure}[t!]
\centering\includegraphics[width=\columnwidth]{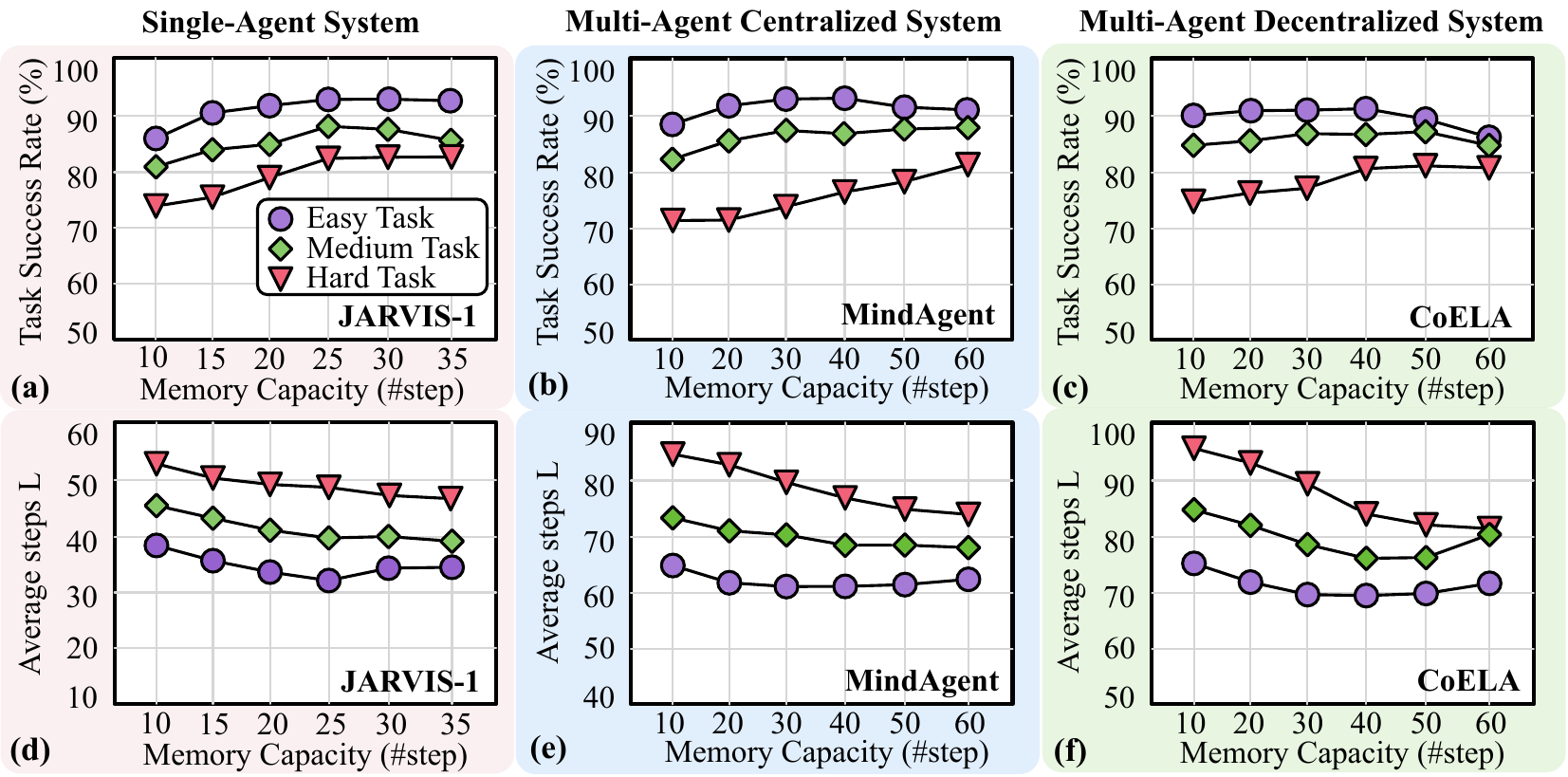}
\vspace{-10pt}
        \caption{\textbf{Memory Module Capacity Analysis.} Average success rate and the number of steps taken to complete the long-horizon tasks across three embodied systems and various memory module capacity.}
        \label{fig:profiling_memory}
        \vspace{-10pt}
\end{figure}

\textbf{Memory module capacity impact.} Fig.~\ref{fig:profiling_memory} illustrates the impact of memory capacity, defined as the amount of past step information stored, on task success rate and completion steps. We observe that \emph{increasing memory capacity generally enhances task performance.} As memory capacity grows, success rates improve while the number of steps decreases. Simpler tasks achieve high success rates with smaller memory sizes, whereas more complex tasks benefit from larger memory modules. However, increasing memory capacity also leads to longer information retrieval times per step.

\textbf{Memory inconsistency.} Fig.~\ref{fig:profiling_memory} also highlights that \emph{memory inconsistency issues arise with excessively large memory capacities}. Performance declines slightly when the memory size becomes very large (e.g., full state-action history). We hypothesize this occurs because, as in-context examples grow, LLMs struggle to retain key details, such as previous actions and object locations, leading to inconsistencies. Additionally, we observe that retrieval based on multimodal states (vision observations, symbolic information, action history) outperforms approaches that rely solely on text embeddings.

\colorbox{lightgray!50}{\textbf{Takeaway 4:}} \textit{Increasing memory module capacity improves success rates and reduces steps, especially for complex tasks. However, excessively large memory introduces inconsistencies and increases retrieval time per step.}

\colorbox{brown!30}{\textbf{Recommendation 5:}} \textit{The memory module overhead and inconsistency can be optimized with a dual memory structure: long-term memory stores static environmental information, while short-term memory captures real-time updates on agent status, task progress, and recent interactions.}

\subsection{Communication Module: Token Length Analysis}
\label{subsec:profiling_prompt_length}

\begin{figure}[t!]
\centering\includegraphics[width=\columnwidth]{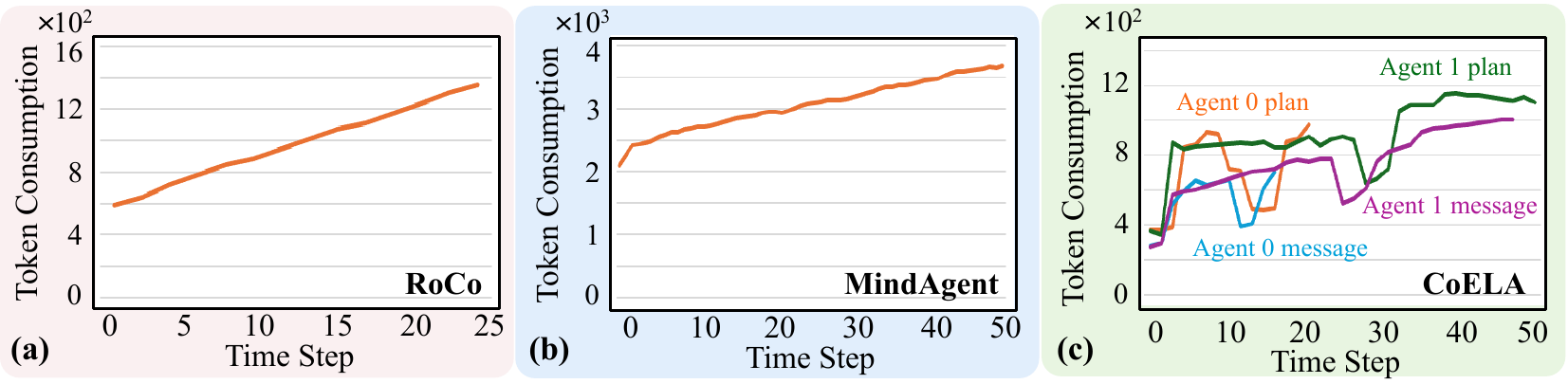}
\vspace{-10pt}
        \caption{\textbf{Prompt Token Length Analysis.} Token length of LLM-based modules over time step across three embodied AI agent systems.}
        \label{fig:profiling_token}
        \vspace{-10pt}
\end{figure}

\textbf{Prompt token length.}
Fig.~\ref{fig:profiling_token} illustrates how prompt token length evolves over time across workloads. We observe that \emph{token length increases as tasks progress, primarily due to input tokens.} As tasks advance, the LLM-based planning and communication modules retrieve more information, often repeating relevant events, which can excessively lengthen prompts and occasionally exceed LLM's token limit. This not only raises computational costs but also reduces LLM's ability to focus on key details, leading to suboptimal responses. In multi-agent systems, dialogue from previous agents is concatenated into prompts for subsequent agents, causing token length to grow within each iteration and as the task proceeds.


\colorbox{lightgray!50}{\textbf{Takeaway 5:}} \textit{Prompt token length increases as tasks progress, driven by repeated information retrieval and concatenated multi-agent dialogues, leading to higher computational costs, which can degrade system efficiency.}

\colorbox{brown!30}{\textbf{Recommendation 6:}} \textit{Prompt length inefficiency can be optimized through context-aware management and compression techniques, such as summarizing dialogue history, removing irrelevant information, and compressing repeated patterns to keep the LLM context both efficient and relevant.}

\subsection{Modular Embodied System: Pipeline Efficiency Analysis}
\label{subsec:profiling_system_inefficiency}
\textbf{System inefficiency.}
Through latency analysis (Fig.~\ref{fig:profiling_latency_breakdown}), we also observe several causes of system-level inefficiencies.
\emph{Sequential processing leads to high system latency.} Specifically, the perception-planning-communication-execution pipeline introduces cumulative delays at each step, along with potential redundancies in high-level planning computations.
Moreover, \emph{inefficient communication and planning mechanisms result in unnecessary dialogues.} 
We discover frequent occasions where agents send redundant, repetitive messages and interfere with one another.
For example, in CoELA, communication is executed before planning, with messages pre-generated at every step for each agent. However, we find that only 20\% of these steps lead to actual communication after agents finalize their plans. The majority of messages are unnecessary and do not contribute to task success yet still add to task latency.


\colorbox{lightgray!50}{\textbf{Takeaway 6:}} \textit{Sequential processing within the modular pipeline and across action steps leads to cumulative delays and redundant high-level planning computations. Inefficient communication mechanisms, such as pre-generating unnecessary messages, hinder effective cooperation and increase latency.}

\colorbox{brown!30}{\textbf{Recommendation 7:}} \textit{The sequential processing can be optimized by planning-guided multi-step execution. Instead of generating a new high-level plan for each low-level action, the planning module can produce a high-level plan that guides multiple consecutive low-level actions over a defined period.}

\colorbox{brown!30}{\textbf{Recommendation 8:}} \textit{The redundant communication can be optimized by planning-then-communication strategy, where the planning module first determines if communication is necessary, only when deemed essential does the system initiate message generation through the LLM. Additionally, hierarchically structuring communication between agents can further enhance effectiveness and improve overall system efficiency.}

\begin{figure}[t!]
\centering\includegraphics[width=\columnwidth]{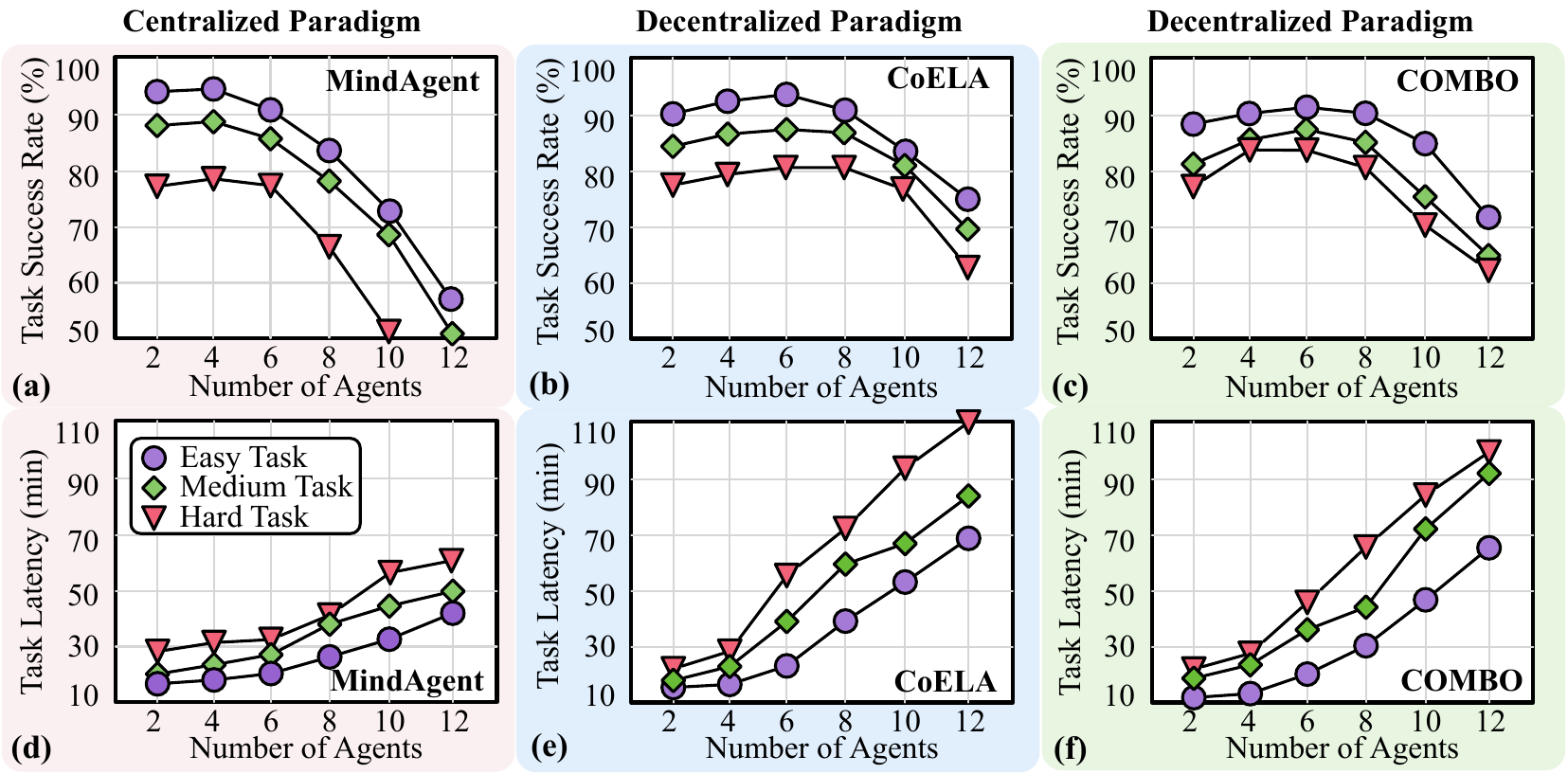}
\vspace{-10pt}
        \caption{\textbf{Multi-Agent System Scalability Analysis.} Average task success rate and end-to-end latency of centralized (MindAgent~\cite{gong2023mindagent}) and decentralized (CoELA~\cite{zhang2024building} and COMBO~\cite{zhang2024combo}) embodied systems across varying number of agents and task difficulty levels.}
        \label{fig:profiling_scalability}
        \vspace{-10pt}
\end{figure}




\section{Embodied System Scalability Analysis}
\label{sec:profiling_scalability}

This section analyzes the scalability of embodied systems. Most multi-agent studies are constrained to 2-4 agents, but scaling to larger-scale systems presents significant challenges, especially in complex long-horizon multi-objective tasks.


\textbf{Scalability challenges.} We identify three key scalability challenges in multi-agent systems.
\emph{First}, the number of possible coordinated actions and their interdependencies grows exponentially with the number of agents, making LLM reasoning increasingly complex.
\emph{Second}, the LLM context includes not only current-round responses from other agents but also dialogue history, actions, and states from previous rounds. As the number of agents grows, the context length expands, approaching LLM token limits and increasing both inference latency and API costs.
\emph{Third}, longer prompts tend to dilute relevant information, further degrading task performance.

Fig.~\ref{fig:profiling_scalability} illustrates the average task success rate and end-to-end latency for centralized (MindAgent) and decentralized (CoELA and COMBO) embodied systems across various number of agents and task difficulty levels. We observe distinct scalability characteristics of these two paradigms in terms of task performance and system efficiency.

\textbf{Centralized system scalability analysis.} 
\emph{Task performance in centralized systems declines sharply with more agents.} As shown in Fig.~\ref{fig:profiling_scalability}\blue{a}, as the number of agents increases, the task success rate drops significantly, indicating that a single central LLM planner struggles to generate effective plans for complex reasoning tasks, resulting in suboptimal performance in intricate scenarios.
On the other hand, as shown in Fig.~\ref{fig:profiling_scalability}\blue{d}, \emph{centralized systems demonstrate better scalability in system efficiency.} This is primarily because they require fewer LLM runs (API calls) and tokens, with these computational requirements scaling linearly as the number of agents increases.

\textbf{Decentralized system scalability analysis.} \emph{Decentralized embodied systems suffer from limited scalability and exploded latency.} As shown in Fig.~\ref{fig:profiling_scalability}\blue{e}-\ref{fig:profiling_scalability}\blue{f}, with more agents, the number of communication rounds per planning step grows, often resulting in repetitive and unproductive dialogues, leading to significant latency. Agents frequently reiterate prior suggestions or propose identical actions, which dilutes the context and hampers collaboration. Additionally, decentralized systems require more LLM runs (API calls) and tokens, with computational demands scaling quadratically with the number of agents. 
In terms of task performance, as the number of agents increases, task success initially improves but declines due to reduced collaboration efficiency in larger agent groups (Fig.~\ref{fig:profiling_scalability}\blue{b}-\ref{fig:profiling_scalability}\blue{c}). 



\colorbox{lightgray!50}{\textbf{Takeaway 7:}} \textit{Multi-agent embodied systems face scalability challenges in long-horizon tasks as the number of agents increases. Centralized systems struggle with performance drops due to the central planner's difficulty in managing complex reasoning, while decentralized systems face inefficiencies from redundant and unproductive dialogues and reduced collaboration efficiency in large teams.}

\colorbox{brown!30}{\textbf{Recommendation 9:}} \textit{The scalability challenges of multi-agent embodied systems can potentially be optimized through a hierarchical cooperative paradigm. Agents are grouped into clusters when close enough to interact, cooperating centrally within clusters and decentrally across clusters.}

\colorbox{brown!30}{\textbf{Recommendation 10:}} \textit{Optimizing communication protocols (e.g., message filtering, prioritization, on-demand generation) can minimize unproductive communication and latency. Decomposing complex tasks into smaller modular subtasks can enable parallel execution and reduce computational burden.}




\section{Related Work}
\label{sec:related_work}

\textbf{Embodied AI Systems.} Recent studies have demonstrated that embodied AI systems, empowered by the advanced reasoning capabilities of LLMs, achieve remarkable performance across a range of complex tasks~\cite{duan2022survey, liu2024aligningcyberspacephysical, xu2024survey}. These systems can be broadly categorized into end-to-end models~\cite{brohan2023rt, li2024generalistrobotpoliciesmatters, hu2023gaia, zhen20243d, team2024octo}, single-agent systems~\cite{chen2023robogpt, wang2024mobile, qin2024mp5, song2023llm}, and multi-agent systems. The multi-agent category can be further subdivided into centralized systems~\cite{yu2023co, zhou2024algpt, gong2023mindagent, liu2024coherent} and decentralized systems~\cite{mandi2024roco, zhang2024combo, chen2023agentverse, zhang2024building}, based on their communication paradigms. Despite their recent success, experimental evidence reveals that many of these architectures suffer from substantial end-to-end latency. 
\emph{To address this gap, this paper presents the in-depth characterization of embodied AI systems, aiming to enable more efficient and scalable execution and deployment.}

\textbf{Emerging Workload Benchmark.} Benchmark and workload characterization are crucial for researchers to understand and evaluate their proposed methods for future optimization. Beyond DNNs and LLMs, extensive studies have been conducted in emerging fields such as  neuromorphic AI~\cite{yik2023neurobench,chang202373,wu2022ubrain}, mixed-reality~\cite{kwon2023xrbench,li2024gaze}, UAVs~\cite{krishnan2022roofline,liu2022energy,krishnan2022automatic,wan2024mulberry}, robotics~\cite{neuman2021robomorphic,wan2021energy,neuman2023roboshape,mayoral2024robotperf,hao2024orianna}, cognitive systems~\cite{wan2024towards,wan2024towards_2,ibrahim2024special,wan2025cogsys}, genome sequencing~\cite{fujiki2018genax,fujiki2020seedex}, graph analytics~\cite{gao2023mega}, mobile systems~\cite{liu2020systolic,liu2022s2ta}, and so on. \emph{While Embodied AI agents demonstrate strong potential in tackling complex tasks, they often face significant inefficiencies and lack thorough system-level analysis. A deeper understanding of their architectural and computational characteristics is therefore essential for optimizing performance.}
\section{Discussion and Conclusion}
\label{sec:conclusion}

Optimizing embodied AI systems requires addressing both intra- and inter-module inefficiencies. Intra-module optimizations focus on enhancing the efficiency of individual components, such as deploying lightweight LLMs, refining low-level control mechanisms, optimizing agent dialogue prompts, and constraining the action space. Leveraging a dual memory system that integrates long-term and short-term memory improves retrieval efficiency and minimizes inconsistencies.

Inter-module optimizations focus on collaborative strategies: for single-agent systems, planning-guided multi-step execution reduces unnecessary planning overhead, while multi-agent systems benefit from a planning-then-communication approach to eliminate redundant dialogues, hierarchical paradigms and task decomposition for efficient collaboration.

These optimizations collectively position embodied AI systems as a transformative paradigm for next-generation intelligent, efficient, and trustworthy autonomous agents. This paper systematically categorizes embodied system paradigms, benchmarks their performance, proposes optimization techniques, and explores opportunities for advancing future embodied autonomous systems.


 
\section*{Acknowledgments}
We thank the anonymous ISPASS reviewers for their comments and feedback. This work was supported in part by CoCoSys, one of the seven centers in JUMP 2.0, a Semiconductor Research Corporation (SRC) program sponsored by DARPA, and NSF GRFP DGE-214074.

\bibliographystyle{ieeetr}
\bibliography{refs}

\end{document}